# An optimization algorithm inspired by the States of Matter that improves the balance between exploration and exploitation

[a]Erik Cuevas[1], [a]Alonso Echavarría, [b]Marte A. Ramírez-Ortegón

[a]Departamento de Electrónica
Universidad de Guadalajara, CUCEI
Av. Revolución 1500, Guadalajara, Jal, México
[1]erik.cuevas@cucei.udg.mx

[b]Institut für Nachrichtentechnik
Technische Universität Braunschweig
Schleinitzstrae 22, 38106 Braunschweig, Germany

**Abstract**

The ability of an Evolutionary Algorithm (EA) to find a global optimal solution depends on its capacity to find a good rate between exploitation of found-so-far elements and exploration of the search space. Inspired by natural phenomena, researchers have developed many successful evolutionary algorithms which, at original versions, define operators that mimic the way nature solves complex problems, with no actual consideration of the exploration-exploitation balance. In this paper, a novel nature-inspired algorithm called the States of Matter Search (SMS) is introduced. The SMS algorithm is based on the simulation of the states of matter phenomenon. In SMS, individuals emulate molecules which interact to each other by using evolutionary operations which are based on the physical principles of the thermal-energy motion mechanism. The algorithm is devised by considering each state of matter at one different exploration–exploitation ratio. The evolutionary process is divided into three phases which emulate the three states of matter: gas, liquid and solid. In each state, molecules (individuals) exhibit different movement capacities. Beginning from the gas state (pure exploration), the algorithm modifies the intensities of exploration and exploitation until the solid state (pure exploitation) is reached. As a result, the approach can substantially improve the balance between exploration–exploitation, yet preserving the good search capabilities of an evolutionary approach. To illustrate the proficiency and robustness of the proposed algorithm, it is compared to other well-known evolutionary methods including novel variants that incorporate diversity preservation schemes. The comparison examines several standard benchmark functions which are commonly considered within the EA field. Experimental results show that the proposed method achieves a good performance in comparison to its counterparts as a consequence of its better exploration–exploitation balance.

*Keywords:* Evolutionary algorithms, Global Optimization, Nature-inspired algorithms.

## 1. Introduction

Global optimization [1] has delivered applications for many areas of science, engineering, economics and others, where mathematical modelling is used [2]. In general, the goal is to find a global optimum for an objective function which is defined over a given search space. Global optimization algorithms are usually broadly divided into deterministic and stochastic [3]. Since deterministic methods only provide a theoretical guarantee of locating a local minimum of the objective function, they often face great difficulties in solving global optimization problems [4]. On the other hand, evolutionary algorithms are usually faster in locating a global optimum [5]. Moreover, stochastic methods adapt easily to black-box formulations and extremely ill-behaved functions, whereas deterministic methods usually rest on at least some theoretical assumptions about the problem formulation and its analytical properties (such as Lipschitz continuity) [6].

Evolutionary algorithms, which are considered as members of the stochastic group, have been developed by a combination of rules and randomness that mimics several natural phenomena. Such phenomena include evolutionary processes such as the Evolutionary Algorithm (EA) proposed by Fogel et al. [7], De Jong [8], and Koza [9], the Genetic Algorithm (GA) proposed by Holland [10] and Goldberg [11], the Artificial Immune System proposed by De Castro et al. [12] and the Differential Evolution Algorithm (DE) proposed by Price & Storn [13]. Some other methods which are based on physical processes include the Simulated Annealing proposed by Kirkpatrick et al. [14], the Electromagnetism-like Algorithm proposed by İlker et al. [15] and the Gravitational Search Algorithm proposed by Rashedi et al. [16]. Also, there are other methods based on the animal-behavior phenomena such as the Particle Swarm Optimization (PSO) algorithm proposed by Kennedy & Eberhart [17] and the Ant Colony Optimization (ACO) algorithm proposed by Dorigo et al. [18].

---

[1] Corresponding author, Tel +52 33 1378 5900, ext. 27714, E-mail: erik.cuevas@cucei.udg.mx

Every EA needs to address the issue of exploration-exploitation of the search space. Exploration is the process of visiting entirely new points of a search space whilst exploitation is the process of refining those points within the neighborhood of previously visited locations, in order to improve their solution quality. Pure exploration degrades the precision of the evolutionary process but increases its capacity to find new potential solutions. On the other hand, pure exploitation allows refining existent solutions but adversely driving the process to local optimal solutions. Therefore, the ability of an EA to find a global optimal solution depends on its capacity to find a good balance between the exploitation of found-so-far elements and the exploration of the search space [19]. So far, the exploration–exploitation dilemma has been an unsolved issue within the framework of EA.

Although PSO, DE and GSA are considered the most popular algorithms for many optimization applications, they fail in finding a balance between exploration and exploitation [20]; in multimodal functions, they do not explore the whole region effectively and often suffers premature convergence or loss of diversity. In order to deal with this problem, several proposals have been suggested in the literature [21-46]. In most of the approaches, exploration and exploitation is modified by the proper settings of control parameters that have an influence on the algorithm's search capabilities [47]. One common strategy is that EAs should start with exploration and then gradually change into exploitation [48]. Such a policy can be easily described with deterministic approaches where the operator that controls the individual diversity decreases along with the evolution. This is generally correct, but such a policy tends to face difficulties when solving certain problems with multimodal functions that hold many optima, since a premature takeover of exploitation over exploration occurs. Some approaches that use this strategy can be found in [21-29]. Other works [30-34] use the population size as reference to change the balance between exploration and exploitation. A larger population size implies a wider exploration while a smaller population demands a shorter search. Although this technique delivers an easier way to keep diversity, it often represents an unsatisfactory solution. An improper handling of large populations might converge to only one point, despite introducing more function evaluations. Recently, new operators have been added to several traditional evolutionary algorithms in order to improve their original exploration-exploitation capability. Such operators diversify particles whenever they concentrate on a local optimum. Some methods that employ this technique are discussed in [35-46].

Either of these approaches is necessary but not sufficient to tackle the problem of the exploration–exploitation balance. Modifying the control parameters during the evolution process without the incorporation of new operators to improve the population diversity, makes the algorithm defenseless against the premature convergence and may result in poor exploratory characteristics of the algorithm [48]. On the other hand, incorporating new operators without modifying the control parameters leads to increase the computational cost, weakening the exploitation process of candidate regions [39]. Therefore, it does seem reasonable to incorporate both of these approaches into a single algorithm.

In this paper, a novel nature-inspired algorithm, known as the States of Matter Search (SMS) is proposed for solving global optimization problems. The SMS algorithm is based on the simulation of the states of matter phenomenon. In SMS, individuals emulate molecules which interact to each other by using evolutionary operations based on the physical principles of the thermal-energy motion mechanism. Such operations allow the increase of the population diversity and avoid the concentration of particles within a local minimum. The proposed approach combines the use of the defined operators with a control strategy that modifies the parameter setting of each operation during the evolution process. In contrast to other approaches that enhance traditional EA algorithms by incorporating some procedures for balancing the exploration–exploitation rate, the proposed algorithm naturally delivers such property as a result of mimicking the states of matter phenomenon. The algorithm is devised by considering each state of matter at one different exploration–exploitation ratio. Thus, the evolutionary process is divided into three stages which emulate the three states of matter: gas, liquid and solid. At each state, molecules (individuals) exhibit different behaviors. Beginning from the gas state (pure exploration), the algorithm modifies the intensities of exploration and exploitation until the solid state (pure exploitation) is reached. As a result, the approach can substantially improve the balance between exploration–exploitation, yet preserving the good search capabilities of an evolutionary approach. To illustrate the proficiency and robustness of the proposed algorithm, it has been compared to other well-known evolutionary methods including recent variants that incorporate diversity preservation schemes. The comparison examines several standard benchmark functions which are usually employed within the EA field. Experimental results show that the proposed method achieves good performance over its counterparts as a consequence of its better exploration–exploitation capability.

This paper is organized as follows. Section 2 introduces basic characteristics of the three states of matter. In Section 3, the novel SMS algorithm and its characteristics are both described. Section 4 presents experimental results and a comparative study. Finally, in Section 5, some conclusions are discussed.

## 2. States of matter

The matter can take different phases which are commonly known as states. Traditionally, three states of matter are known: solid, liquid, and gas. The differences among such states are based on forces which are exerted among particles composing a material [49].

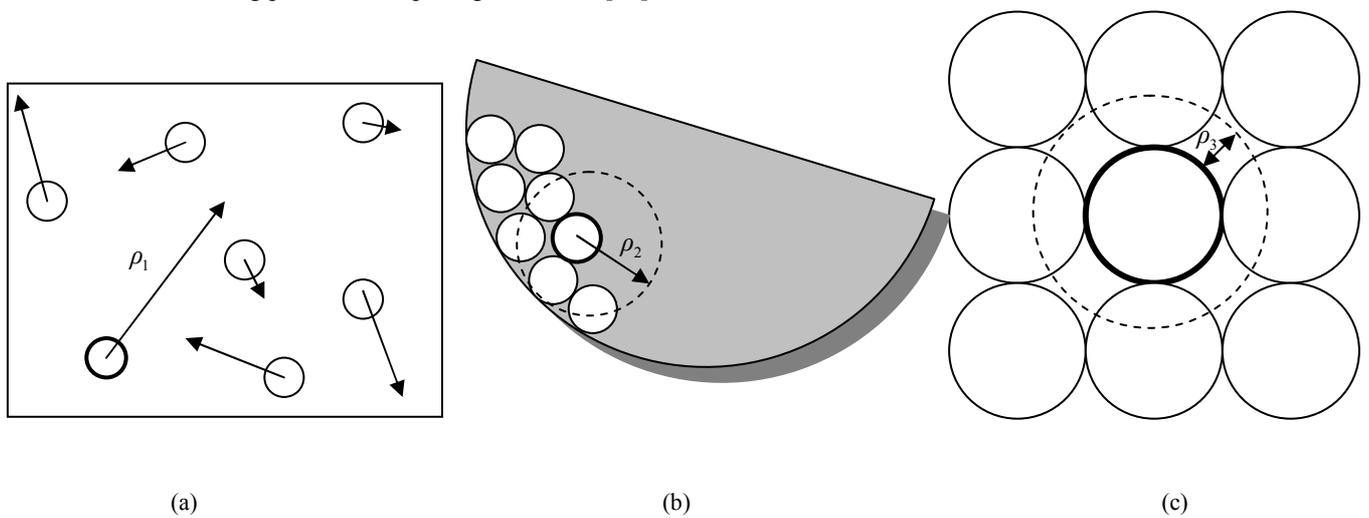

**Fig. 1.** Different states of matter: (a) gas, (b) liquid, and (c) solid.

In the gas phase, molecules present enough kinetic energy so that the effect of intermolecular forces is small (or zero for an ideal gas), while the typical distance between neighboring molecules is greater than the molecular size. A gas has no definite shape or volume, but occupies the entire container in which it is confined. Fig. 1a shows the movements exerted by particles in a gas state. The movement experimented by the molecules represent the maximum permissible displacement $\rho_1$ among particles [50]. In a liquid state, intermolecular forces are more restrictive than those in the gas state. The molecules have enough energy to move relatively to each other still keeping a mobile structure. Therefore, the shape of a liquid is not definite but is determined by its container. Fig. 1b presents a particle movement $\rho_2$ within a liquid state. Such movement is smaller than those considered by the gas state but larger than the solid state [51]. In the solid state, particles (or molecules) are packed together closely with forces among particles being strong enough so that the particles cannot move freely but only vibrate. As a result, a solid has a stable, definite shape and a definite volume. Solids can only change their shape by force, as when they are broken or cut. Fig. 1c shows a molecule configuration in a solid state. Under such conditions, particles are able to vibrate (being perturbed) considering a minimal $\rho_3$ distance [50].

In this paper, a novel nature-inspired algorithm known as the States of Matter Search (SMS) is proposed for solving global optimization problems. The SMS algorithm is based on the simulation of the states of matter phenomenon that considers individuals as molecules which interact to each other by using evolutionary operations based on the physical principles of the thermal-energy motion mechanism. The algorithm is devised by considering each state of matter at one different exploration–exploitation ratio. Thus, the evolutionary process is divided into three stages which emulate the three states of matter: gas, liquid and solid. In each state, individuals exhibit different behaviors.

## 3. States of matter search (SMS)

*3.1 Definition of Operators*

In the approach, individuals are considered as molecules whose positions on a multidimensional space are modified as the algorithm evolves. The movement of such molecules is motivated by the analogy to the motion of thermal-energy.

The velocity and direction of each molecule's movement are determined by considering the collision, the attraction forces and the random phenomena experimented by the molecule set [52]. In our approach, such behaviors have been implemented by defining several operators such as the direction vector, the collision and the random positions operators, all of which emulate the behavior of actual physics laws.

The direction vector operator assigns a direction to each molecule in order to lead the particle movement as the evolution process takes place. On the other side, the collision operator mimics those collisions that are experimented by molecules as they interact to each other. A collision is considered when the distance between two molecules is shorter than a determined proximity distance. The collision operator is thus implemented by interchanging directions of the involved molecules. In order to simulate the random behavior of molecules, the proposed algorithm generates random positions following a probabilistic criterion that considers random locations within a feasible search space.

The next section presents all operators that are used in the algorithm. Although such operators are the same for all the states of matter, they are employed over a different configuration set depending on the particular state under consideration.

### 3.1.1 Direction vector

The direction vector operator mimics the way in which molecules change their positions as the evolution process develops. For each *n*-dimensional molecule $\mathbf{p}_i$ from the population $\mathbf{P}$, it is assigned an *n*-dimensional direction vector $\mathbf{d}_i$ which stores the vector that controls the particle movement. Initially, all the direction vectors ($\mathbf{D} = \{\mathbf{d}_1, \mathbf{d}_2, \ldots, \mathbf{d}_{N_p}\}$) are randomly chosen within the range of [-1,1].

As the system evolves, molecules experiment several attraction forces. In order to simulate such forces, the proposed algorithm implements the attraction phenomenon by moving each molecule towards the best so-far particle. Therefore, the new direction vector for each molecule is iteratively computed considering the following model:

$$\mathbf{d}_i^{k+1} = \mathbf{d}_i^k \cdot \left(1 - \frac{k}{gen}\right) \cdot 0.5 + \mathbf{a}_i, \qquad (1)$$

where $\mathbf{a}_i$ represents the attraction unitary vector calculated as $\mathbf{a}_i = (\mathbf{p}^{best} - \mathbf{p}_i) / \|\mathbf{p}^{best} - \mathbf{p}_i\|$, being $\mathbf{p}^{best}$ the best individual seen so-far, while $\mathbf{p}_i$ is the molecule *i* of population $\mathbf{P}$. *k* represents the iteration number whereas *gen* involves the total iteration number that constitutes the complete evolution process.

Under this operation, each particle is moved towards a new direction which combines the past direction, which was initially computed, with the attraction vector over the best individual seen so-far. It is important to point out that the relative importance of the past direction decreases as the evolving process advances. This particular type of interaction avoids the quick concentration of information among particles and encourages each particle to search around a local candidate region in its neighborhood, rather than enteracting to a particle lying at distant region of the domain. The use of this scheme has two advantages: first, it prevents the particles from moving toward the global best position in early stages of algorithm and thus makes the algorithm less susceptible to premature convergence; second, it encourages particles to explore their own neighborhood thoroughly, just before they converge towards a global best position. Therefore, it provides the algorithm with local search ability enhancing the exploitative behavior.

In order to calculate the new molecule position, it is necessary to compute the velocity $\mathbf{v}_i$ of each molecule by using:

$$\mathbf{v}_i = \mathbf{d}_i \cdot v_{init} \tag{2}$$

being $v_{init}$ the initial velocity magnitude which is calculated as follows:

$$v_{init} = \frac{\sum_{j=1}^{n}(b_j^{high} - b_j^{low})}{n} \cdot \beta \tag{3}$$

where $b_j^{low}$ and $b_j^{high}$ are the low $j$ parameter bound and the upper $j$ parameter bound respectively, whereas $\beta \in [0,1]$.

Then, the new position for each molecule is updated by:

$$p_{i,j}^{k+1} = p_{i,j}^{k} + v_{i,j} \cdot \text{rand}(0,1) \cdot \rho \cdot (b_j^{high} - b_j^{low}) \tag{4}$$

where $0.5 \leq \rho \leq 1$.

3.1.2 Collision

The collision operator mimics the collisions experimented by molecules while they interact to each other. Collisions are calculated if the distance between two molecules is shorter than a determined proximity value. Therefore, if $\|\mathbf{p}_i - \mathbf{p}_q\| < r$, a collision between molecules $i$ and $q$ is assumed; otherwise, there is no collision, considering $i, q \in \{1, \ldots, N_p\}$ such that $i \neq q$. If a collision occurs, the direction vector for each particle is modified by interchanging their respective direction vectors as follows:

$$\mathbf{d}_i = \mathbf{d}_q \text{ and } \mathbf{d}_q = \mathbf{d}_i \tag{5}$$

The collision radius is calculated by:

$$r = \frac{\sum_{j=1}^{n}(b_j^{high} - b_j^{low})}{n} \cdot \alpha \tag{6}$$

where $\alpha \in [0,1]$.

Under this operator, a spatial region enclosed within the radius $r$ is assigned to each particle. In case the particle regions collide to each other, the collision operator acts upon particles by forcing them out of the region. The radio $r$ and the collision operator provide the ability to control diversity throughout the search process. In other words, the rate of increase or decrease of diversity is predetermined for each stage. Unlike other diversity-guided algorithms, it is not necessary to inject diversity into the population when particles gather around a local optimum because the diversity will be preserved during the overall search process. The collision incorporation therefore enhances the exploratory behavior in the proposed approach.

### 3.1.3 Random positions

In order to simulate the random behavior of molecules, the proposed algorithm generates random positions following a probabilistic criterion within a feasible search space.

For this operation, a uniform random number $r_m$ is generated within the range [0,1]. If $r_m$ is smaller than a threshold $H$, a random molecule's position is generated; otherwise, the element remains with no change. Therefore such operation can be modeled as follows:

$$p_{i,j}^{k+1} = \begin{cases} b_j^{low} + \text{rand}(0,1) \cdot (b_j^{high} - b_j^{low}) & \text{with probability } H \\ p_{i,j}^{k+1} & \text{with probability } (1-H) \end{cases} \quad (7)$$

where $i \in \{1,\ldots,N_p\}$ and $j \in \{1,\ldots,n\}$.

### 3.1.4 Best Element Updating

Despite this updating operator does not belong to State of Matter metaphor, it is used to simply store the best so-far solution. In order to update the best molecule $\mathbf{p}^{best}$ seen so-far, the best found individual from the current $k$ population $\mathbf{p}^{best,k}$ is compared to the best individual $\mathbf{p}^{best,k-1}$ of the last generation. If $\mathbf{p}^{best,k}$ is better than $\mathbf{p}^{best,k-1}$ according to its fitness value, $\mathbf{p}^{best}$ is updated with $\mathbf{p}^{best,k}$, otherwise $\mathbf{p}^{best}$ remains with no change. Therefore, $\mathbf{p}^{best}$ stores the best historical individual found so-far.

### *3.2 SMS algorithm*

The overall SMS algorithm is composed of three stages corresponding to the three States of Matter: the gas, the liquid and the solid state. Each stage has its own behavior. In the first stage (gas state), exploration is intensified whereas in the second one (liquid state) a mild transition between exploration and exploitation is executed. Finally, in the third phase (solid state), solutions are refined by emphasizing the exploitation process.

### 3.2.1 General procedure

At each stage, the same operations are implemented. However, depending on which state is referred, they are employed considering a different parameter configuration. The general procedure in each state is shown as pseudo-code in Algorithm 1. Such procedure is composed by five steps and maps the current population $\mathbf{P}^k$ to a new population $\mathbf{P}^{k+1}$. The algorithm receives as input the current population $\mathbf{P}^k$ and the configuration parameters $\rho$, $\beta$, $\alpha$, and $H$, whereas it yields the new population $\mathbf{P}^{k+1}$.

**Step 1:** Find the best element of the population **P**
$$\mathbf{p}^{best} \in \{\mathbf{P}\} \mid f(\mathbf{p}^{best}) = \max\{f(\mathbf{p}_1), f(\mathbf{p}_2),\ldots,f(\mathbf{p}_{N_p})\}$$

**Step 2:** Calculate $v_{init}$ and $r$
$$v_{init} = \frac{\sum_{j=1}^n (b_j^{high} - b_j^{low})}{n} \cdot \beta \qquad r = \frac{\sum_{j=1}^n (b_j^{high} - b_j^{low})}{n} \cdot \alpha$$

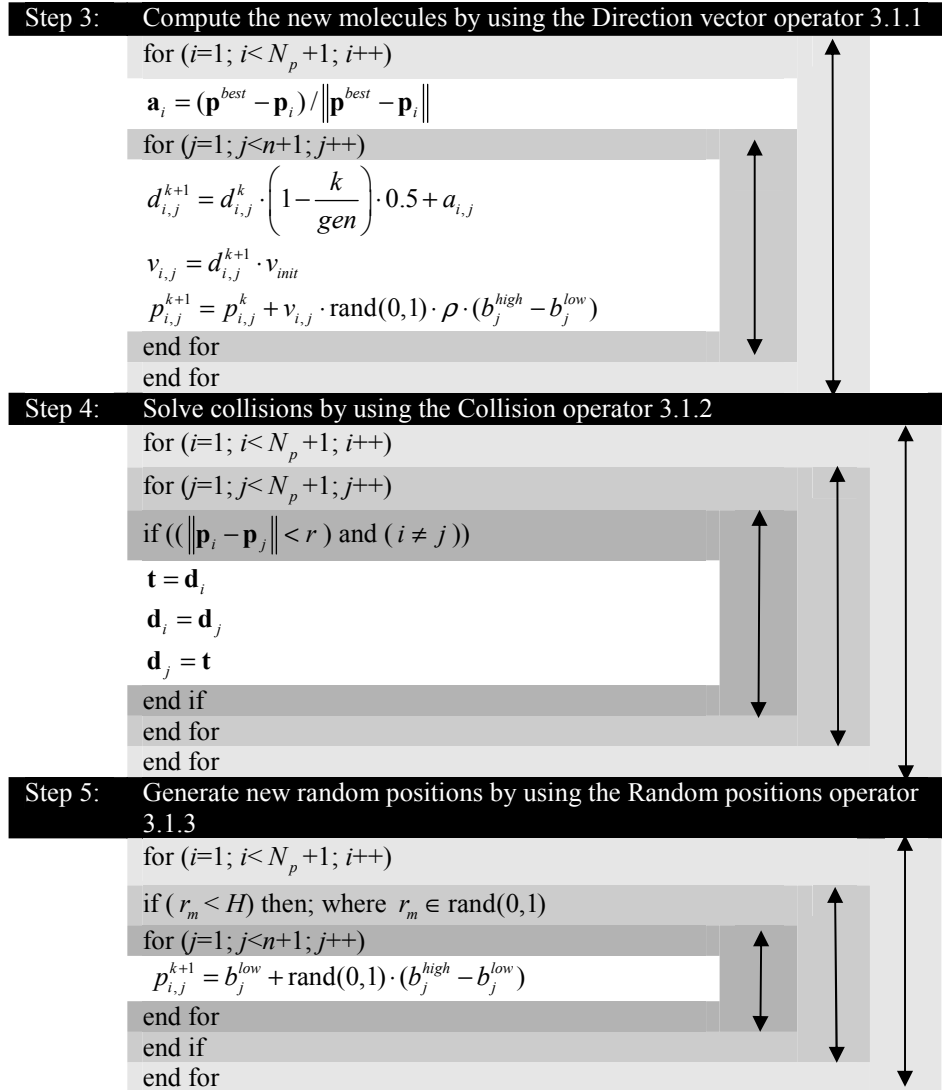

**Algorithm 1.** General procedure executed by all the states of matter.

### 3.2.2 The complete algorithm

The complete algorithm is divided into four different parts. The first corresponds to the initialization stage, whereas the last three represent the States of Matter. All the optimization process, which consists of a *gen* number of iterations, is organized into three different asymmetric phases, employing 50% of all iterations for the gas state (exploration), 40% for the liquid state (exploration-exploitation) and 10% for the solid state (exploitation). The overall process is graphically described by Figure 2. At each state, the same general procedure (see Algorithm 1) is iteratively used considering the particular configuration predefined for each State of Matter. Figure 3 shows the data flow for the complete SMS algorithm.

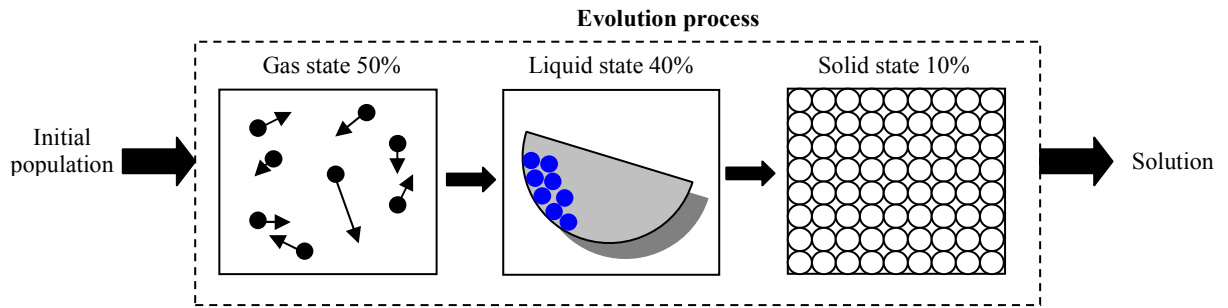

**Fig. 2:** Evolution process in the proposed approach.

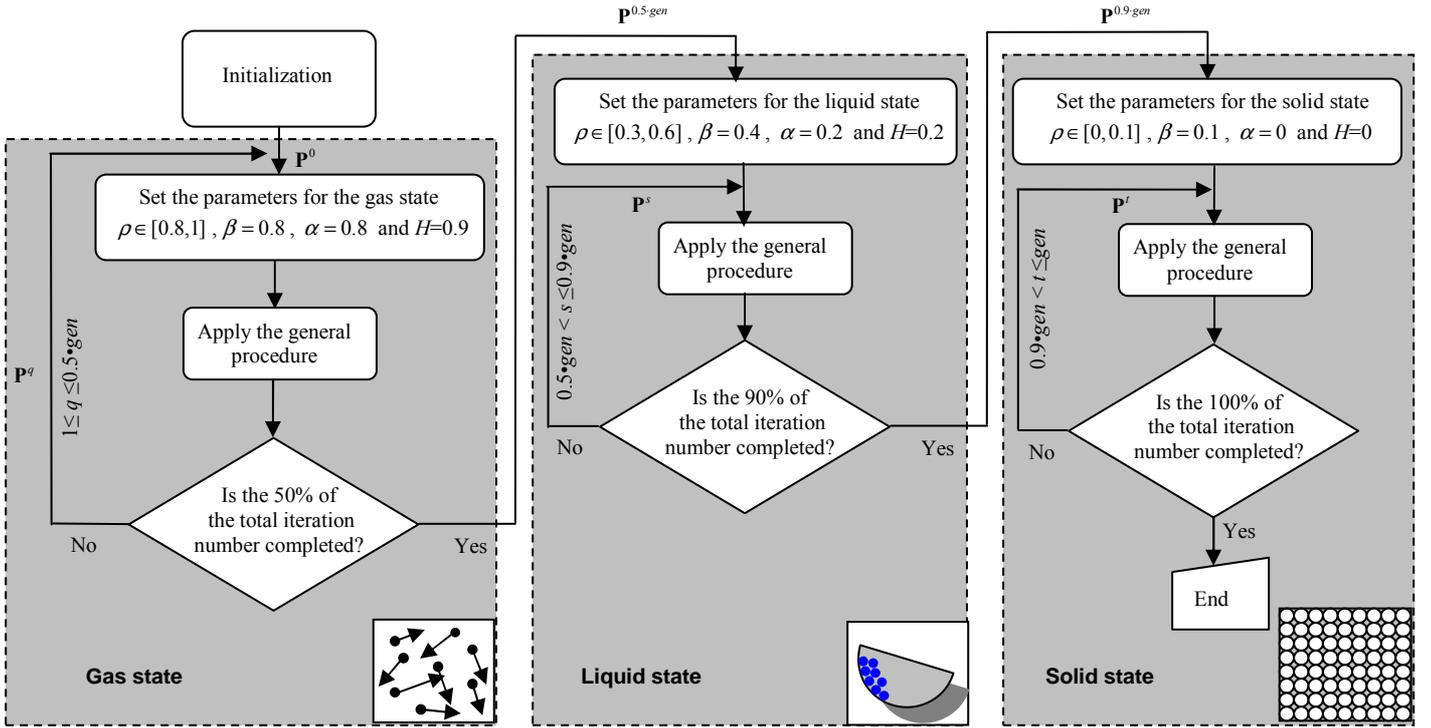

**Fig. 3:** Data flow in the complete SMS algorithm

Initialization

The algorithm begins by initializing a set **P** of $N_p$ molecules ( $\mathbf{P} = \{\mathbf{p}_1, \mathbf{p}_2, \ldots, \mathbf{p}_{N_p}\}$ ). Each molecule position $\mathbf{p}_i$ is a *n*-dimensional vector containing the parameter values to be optimized. Such values are randomly and uniformly distributed between the pre-specified lower initial parameter bound $b_j^{low}$ and the upper initial parameter bound $b_j^{high}$, just as it is described by the following expressions:

$$p_{i,j}^0 = b_j^{low} + \text{rand}(0,1) \cdot (b_j^{high} - b_j^{low}) \qquad (8)$$
$$j = 1, 2, \ldots, n; \quad i = 1, 2, \ldots, N_p,$$

where *j* and *i*, are the parameter and molecule index respectively whereas zero indicates the initial population. Hence, $p_i^j$ is the *j*-th parameter of the *i*-th molecule.

Gas state

In the gas state, molecules experiment severe displacements and collisions. Such state is characterized by random movements produced by non-modeled molecule phenomena [52]. Therefore, the $\rho$ value from the direction vector operator is set to a value near to one so that the molecules can travel longer distances. Similarly, the *H* value representing the random positions operator is also configured to a value around one, in order to allow the random generation for other molecule positions. The gas state is the first phase and lasts for the 50% of all iterations which compose the complete optimization process. The computational procedure for the gas state can be summarized as follows:

Step 1: Set the parameters $\rho \in [0.8, 1]$, $\beta = 0.8$, $\alpha = 0.8$ and *H*=0.9 being consistent with the gas state.
Step 2: Apply the general procedure which is illustrated in Algorithm 1.
Step 3: If the 50% of the total iteration number is completed ($1 \le k \le 0.5 \cdot gen$), then the process continues to the liquid state procedure; otherwise go back to step 2.

Liquid state

Although molecules currently at the liquid state exhibit restricted motion in comparison to the gas state, they still show a higher flexibility with respect to the solid state. Furthermore, the generation of random positions which are produced by non-modeled molecule phenomena is scarce [53]. For this reason, the $\rho$ value from the direction vector operator is bounded to a value between 0.3 and 0.6. Similarly, the random position operator $H$ is configured to a value near to cero in order to allow the random generation of fewer molecule positions. In the liquid state, collisions are also less common than in gas state, so the collision radius, that is controlled by $\alpha$, is set to a smaller value in comparison to the gas state. The liquid state is the second phase and lasts the 40% of all iterations which compose the complete optimization process. The computational procedure for the liquid state can be summarized as follows:

Step 4: Set the parameters $\rho \in [0.3, 0.6]$, $\beta = 0.4$, $\alpha = 0.2$ and $H=0.2$ being consistent with the liquid state.
Step 5: Apply the general procedure that is defined in Algorithm 1.
Step 6: If the 90% (50% from the gas state and 40% from the liquid state) of the total iteration number is completed ($0.5 \cdot gen < k \leq 0.9 \cdot gen$), then the process continues to the solid state procedure; otherwise go back to step 5.

Solid state

In the solid state, forces among particles are stronger so that particles cannot move freely but only vibrate. As a result, effects such as collision and generation of random positions are not considered [52]. Therefore, the $\rho$ value of the direction vector operator is set to a value near to zero indicating that the molecules can only vibrate around their original positions. The solid state is the third phase and lasts for the 10% of all iterations which compose the complete optimization process. The computational procedure for the solid state can be summarized as follows:

Step 7: Set the parameters $\rho \in [0.0, 0.1]$ and $\beta = 0.1$, $\alpha = 0$ and $H=0$ being consistent with the solid state.
Step 8: Apply the general procedure that is defined in Algorithm 1.
Step 9: If the 100% of the total iteration number is completed ($0.9 \cdot gen < k \leq gen$), the process is finished; otherwise go back to step 8.

It is important to clarify that the use of this particular configuration ($\alpha = 0$ and $H=0$) disables the collision and generation of random positions operators which have been illustrated in the general procedure.

4. Experimental results

A comprehensive set of 24 functions, collected from Refs. [54-61], has been used to test the performance of the proposed approach. Tables A1–A4 in the Appendix A present the benchmark functions used in our experimental study. Such functions are classified into four different categories: Unimodal test functions (Table A1), multimodal test functions (Table A2), multimodal test functions with fixed dimensions (Table A3) and functions proposed for the GECCO contest (Table A4). In such tables, $n$ indicates the dimension of the function, $f_{opt}$ the optimum value of the function and $S$ the subset of $R^n$. The function optimum position ($\mathbf{x}_{opt}$) for $f_1$, $f_2$, $f_4$, $f_6$, $f_7$, $f_{10}$, $f_{11}$ and $f_{14}$ is at $\mathbf{x}_{opt} = [0]^n$, for $f_3$, $f_8$ and $f_9$ is at $\mathbf{x}_{opt} = [1]^n$, for $f_5$ is at $\mathbf{x}_{opt} = [420.96]^n$, for $f_{18}$ is at $\mathbf{x}_{opt} = [0]^n$, for $f_{12}$ is at $\mathbf{x}_{opt} = [0.0003075]^n$ and for $f_{13}$ is at $\mathbf{x}_{opt} = [-3.32]^n$. In case of functions contained in Table A4, the $\mathbf{x}^{opt}$ and $f_{opt}$ values have been set to default values which have been obtained from the Matlab© implementation for GECCO competitions, as it is provided in [59]. A detailed description of optimum locations is given in Appendix A.

*4.1 Performance comparison to other meta-heuristic algorithms*

We have applied the SMS algorithm to 24 functions whose results have been compared to those produced by the Gravitational Search Algorithm (GSA) [16], the Particle Swarm Optimization (PSO) method [17] and the Differential Evolution (DE) algorithm [13]. These are considered as the most popular algorithms in many optimization applications. In order to enhance the performance analysis, the PSO algorithm with a territorial diversity-preserving scheme (TPSO) [39] has also been added into the comparisons. TPSO is considered a recent PSO variant that incorporates a diversity preservation scheme in order to improve the balance between exploration and exploitation. In all comparisons, the population has been set to 50. The maximum iteration number for functions in Tables A1, A2 and A4 has been set to 1000 and for functions in Table A3 has been set to 500. Such stop criterion has been selected to maintain compatibility to similar works reported in the literature [4,16].

The parameter setting for each algorithm in the comparison is described as follows:

1. GSA [16]: The parameters are set to $G_o = 100$ and $\alpha = 20$; the total number of iterations is set to 1000 for functions $f_1$ to $f_{11}$ and 500 for functions $f_{12}$ to $f_{14}$. The total number of individuals is set to 50. Such values are the best parameter set for this algorithm according to [16].
2. PSO [17]: The parameters are set to $c_1 = 2$ and $c_2 = 2$; besides, the weight factor decreases linearly from 0.9 to 0.2.
3. DE [13]: The DE/Rand/1 scheme is employed. The crossover probability is set to $CR = 0.9$ and the weight factor is set to $F = 0.8$.
4. TPSO [39]: The parameter $\alpha$ has been set to 0.5. Such value is found to be the best configuration according to [39]. The algorithm has been tuned according to the set of values which have been originally proposed by its own reference.

The experimental comparison between metaheuristic algorithms with respect to SMS has been developed according to the function-type classification as follows:

1. Unimodal test functions (Table A1).
2. Multimodal test functions (Table A2).
3. Multimodal test functions with fixed dimension (Table A3).
4. Test functions from the GECCO contest (Table A4).

Unimodal test functions

This experiment is performed over the functions presented in Table A1. The test compares the SMS to other algorithms such as GSA, PSO, DE and TPSO. The results for 30 runs are reported in Table 1 considering the following performance indexes: the Average Best-so-far (AB) solution, the Median Best-so-far (MB) and the Standard Deviation (SD) of best-so-far solution. The best outcome for each function is boldfaced. According to this table, SMS delivers better results than GSA, PSO, DE and TPSO for all functions. In particular, the test remarks the largest difference in performance which is directly related to a better trade-off between exploration and exploitation. Just as it is illustrated by Figure 4, SMS, DE and GSA have similar convergence rates at finding the optimal minimal, yet faster than PSO and TPSO.

A non-parametric statistical significance proof known as the Wilcoxon's rank sum test for independent samples [62,63] has been conducted over the "average best-so-far" (AB) data of Table 1, with an 5% significance level. Table 2 reports the *p*-values produced by Wilcoxon's test for the pair-wise comparison of the "average best so-far" of four groups. Such groups are formed by SMS vs. GSA, SMS vs. PSO, SMS vs. DE and SMS vs. TPSO. As a null hypothesis, it is assumed that there is no significant difference between mean values of the two algorithms. The alternative hypothesis considers a significant difference between the "average best-so-far" values of both approaches. All *p*-values reported in Table 2 are less than 0.05 (5% significance level) which is a strong evidence against the null hypothesis. Therefore, such evidence indicates that SMS results are statistically significant and that it has not occurred by coincidence (i.e. due to common noise contained in the process).

|   |    | SMS | GSA | PSO | DE | TPSO |
|---|----|-----|-----|-----|----|------|
| $f_1$ | AB | **4.68457E-16** | 1.3296E-05 | 0.873813333 | 0.186584241 | 0,100341256 |
|   | MB | 4.50542E-16 | 7.46803E-06 | 4.48139E-12 | 0.189737658 | 0,101347821 |
|   | SD | 1.23694E-16 | 1.45053E-05 | 4.705628811 | 0.039609704 | 0,002421043 |
| $f_2$ | AB | **0.033116745** | 0.173618066 | 12.83021186 | 54.85755486 | 0.103622066 |
|   | MB | 1.02069E-08 | 0.159932758 | 12.48059177 | 54.59915941 | 0,122230612 |
|   | SD | 0.089017369 | 0,122230612 | 3.633980625 | 4.506836836 | 0,006498124 |
| $f_3$ | AB | **19.64056183** | 32.83253962 | 33399.69716 | 46898.34558 | 21.75247912 |
|   | MB | 26.87914282 | 27.65055745 | 565.0810149 | 43772.19502 | 28.45741892 |
|   | SD | 11.8115879 | 19.11361524 | 43099.34439 | 15697.6366 | 14.56258711 |
| $f_4$ | AB | **8.882513655** | 9.083435186 | 15.05362961 | 12.83391861 | 13.98432748 |
|   | MB | 9.016816582 | 9.150769929 | 13.91301428 | 12.89762202 | 14.01237836 |
|   | SD | 0.442124359 | 0.499181789 | 4.790792877 | 0.542197802 | 1.023476914 |

**Table 1.** Minimization result of benchmark functions of Table A1 with *n*=30. Maximum number of iterations=1000.

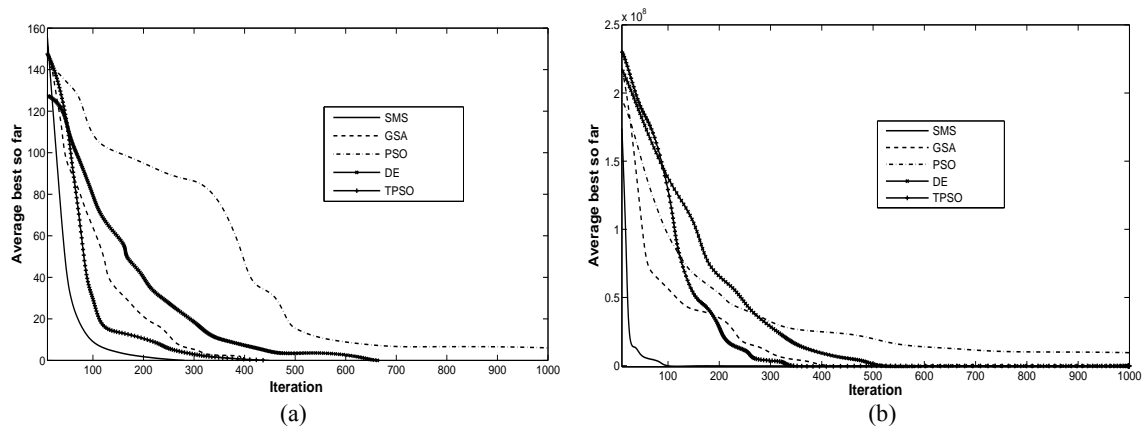

**Fig. 4.** Convergence rate comparison of GSA, PSO, DE, SMS and TPSO for minimization of (a) $f_1$ and (b) $f_3$ considering *n*=30.

| SMS vs | PSO | GSA | DE | TPSO |
|--------|-----|-----|-----|------|
| $f_1$ | $3.94\times10^{-5}$ | $7.39\times10^{-4}$ | $1.04\times10^{-6}$ | $4.12\times10^{-4}$ |
| $f_2$ | $5.62\times10^{-5}$ | $4.92\times10^{-4}$ | $2.21\times10^{-6}$ | $3.78\times10^{-4}$ |
| $f_3$ | $6.42\times10^{-8}$ | $7.11\times10^{-7}$ | $1.02\times10^{-4}$ | $1.57\times10^{-4}$ |
| $f_4$ | $1.91\times10^{-8}$ | $7.39\times10^{-4}$ | $1.27\times10^{-6}$ | $4.22\times10^{-4}$ |

**Table 2.** *p*-values produced by Wilcoxon's test comparing SMS vs. PSO, SMS vs. GSA, SMS vs. DE and SMS vs. TPSO over the "average best-so-far" (AB) values from Table 3.

Multimodal test functions

Multimodal functions represent a good optimization challenge as they possess many local minima (Table A2). In the case of multimodal functions, final results are very important since they reflect the algorithm's ability to escape from poor local optima and to locate a near-global optimum. Experiments using $f_5$ to $f_{11}$ are quite relevant as the number of local minima for such functions increases exponentially as their dimensions increase. The dimension of such functions is set to 30. The results are averaged over 30 runs, reporting the performance index for each function in Table 3 as follows: the Average Best-so-far (AB) solution, the Median Best-so-far (MB) and the Standard Deviation (SD) best-so-far (the best result for each function is highlighted). Likewise, *p*-values of the Wilcoxon signed-rank test of 30 independent runs are listed in Table 4.

| | | SMS | GSA | PSO | DE | TPSO |
|---|---|---|---|---|---|---|
| $f_5$ | AB | **1756.862345** | 9750.440145 | 4329.650468 | 4963.600685 | 1893.673916 |
| | MB | 0.070624076 | 9838.388135 | 4233.282929 | 5000.245932 | 50.23617893 |
| | SD | 1949.048601 | 405.1365297 | 699.7276454 | 202.2888921 | 341.2367823 |
| $f_6$ | AB | **10.95067665** | 15.18970458 | 130.5959941 | 194.6220253 | 18.56962853 |
| | MB | 0.007142491 | 13.9294268 | 129.4942809 | 196.1369499 | 1.234589423 |
| | SD | 14.38387472 | 4.508037915 | 27.87011038 | 9.659933059 | 7.764931264 |
| $f_7$ | AB | **0.000299553** | 0.000575111 | 0.19630233 | 0.98547042 | 0.002348619 |
| | MB | 8.67349E-05 | 0 | 0.011090373 | 0.991214493 | 0.000482084 |
| | SD | 0.000623992 | 0.0021752 | 0.702516846 | 0.031985616 | 0.000196428 |
| $f_8$ | AB | **1.35139E-05** | 2.792846799 | 1450.666769 | 304.6986718 | 1.753493426 |
| | MB | 7.14593E-06 | 2.723230534 | 0.675050254 | 51.86661185 | 1.002364819 |
| | SD | 2.0728E-05 | 1.324814757 | 1708.798785 | 554.2231579 | 0.856294537 |
| $f_9$ | AB | **0.002080591** | 14.49783478 | 136.6888694 | 67251.29956 | 5.284029512 |
| | MB | 0.000675275 | 9.358377669 | 7.00288E-05 | 37143.43153 | 0.934751939 |
| | SD | 0.003150999 | 18.02351657 | 7360.920758 | 63187.52749 | 1.023483601 |
| $f_{10}$ | AB | **0.003412411** | 40.59204902 | 365.7806149 | 822.087914 | 9.636393364 |
| | MB | 0.003164797 | 39.73690704 | 359.104488 | 829.1521586 | 0.362322274 |
| | SD | 0.001997493 | 11.46284891 | 148.9342039 | 81.93476435 | 2.194638533 |
| $f_{11}$ | AB | **0.199873346** | 1.121397135 | 0.857971914 | 3.703467688 | 0.452738336 |
| | MB | 0.199873346 | 1.114194975 | 0.499967033 | 3.729096071 | 0.124948295 |
| | SD | 0.073029674 | 0.271747312 | 1.736399225 | 0.278860779 | 0.247510642 |

**Table 3.** Minimization result of benchmark functions in Table A2 with *n*=30. Maximum number of iterations=1000.

In the case of functions $f_8$, $f_9$, $f_{10}$ and $f_{11}$, SMS yields much better solutions than other methods. However, for functions $f_5$, $f_6$ and $f_7$, SMS produces similar results to GSA and TPSO. The Wilcoxon rank test results, which are presented in Table 4, demonstrate that SMS performed better than GSA, PSO, DE and TPSO considering four functions $f_8 - f_{11}$, whereas, from a statistical viewpoint, there is no difference between results from SMS, GSA and TPSO for $f_5$, $f_6$ and $f_7$. The progress of the "average best-so-far" solution over 30 runs for functions $f_5$ and $f_{11}$ is shown by Fig. 5.

| SMS vs | GSA | PSO | DE | TPSO |
|---|---|---|---|---|
| $f_5$ | 0.087 | $8.38 \times 10^{-4}$ | $4.61 \times 10^{-4}$ | 0.058 |
| $f_6$ | 0.062 | $1.92 \times 10^{-9}$ | $9.97 \times 10^{-8}$ | 0.012 |
| $f_7$ | 0.055 | $4.21 \times 10^{-5}$ | $3.34 \times 10^{-4}$ | 0.061 |
| $f_8$ | $7.74 \times 10^{-9}$ | $3.68 \times 10^{-7}$ | $8.12 \times 10^{-5}$ | $1.07 \times 10^{-5}$ |
| $f_9$ | $1.12 \times 10^{-8}$ | $8.80 \times 10^{-9}$ | $4.02 \times 10^{-8}$ | $9.21 \times 10^{-5}$ |
| $f_{10}$ | $4.72 \times 10^{-9}$ | $3.92 \times 10^{-5}$ | $2.20 \times 10^{-4}$ | $7.41 \times 10^{-5}$ |
| $f_{11}$ | $4.72 \times 10^{-9}$ | $3.92 \times 10^{-5}$ | $2.20 \times 10^{-4}$ | $4.05 \times 10^{-5}$ |

**Table 4.** *p*-values produced by Wilcoxon's test comparing SMS vs. GSA, SMS vs. PSO, SMS vs. DE and SMS vs. TPSO over the "average best-so-far" (AB) values from Table 3

 

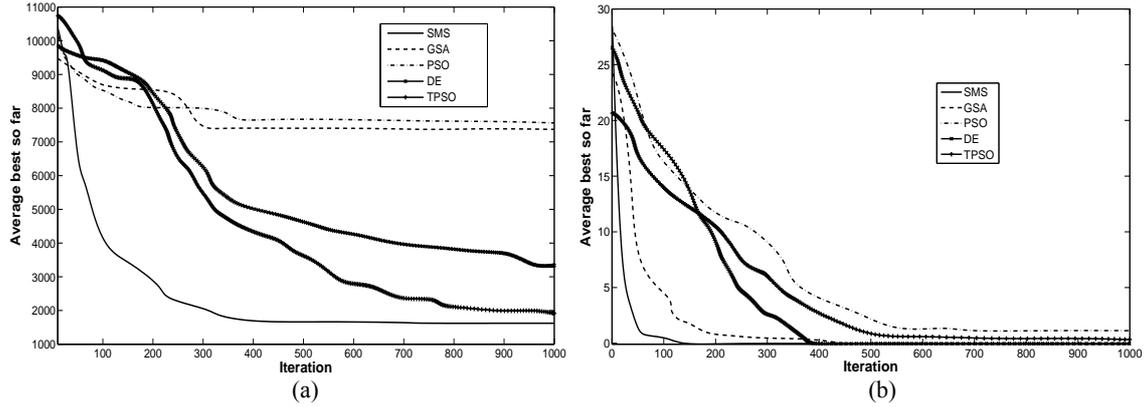

**Fig. 5.** Convergence rate comparison of PSO, GSA, DE, SMS and TPSO for minimization of (a) $f_5$ and (b) $f_{11}$ considering $n=30$.

Multimodal test functions with fixed dimensions

In the following experiments, the SMS algorithm is compared to GSA, PSO, DE and TPSO over a set of multidimensional functions with fixed dimensions which are widely used in the meta-heuristic literature. The functions used for the experiments are $f_{12}$, $f_{13}$ and $f_{14}$ which are presented in Table A3. The results in Table 5 show that SMS, GSA, PSO, DE and TPSO have similar values in their performance. The evidence shows how meta-heuristic algorithms maintain a similar average performance when they face low-dimensional functions [54]. Fig. 6 presents the convergence rate for the GSA, PSO, DE, SMS and TPSO algorithms considering functions $f_{12}$ to $f_{13}$.

|  |  | SMS | GSA | PSO | DE | TPSO |
|---|---|---|---|---|---|---|
| $f_{12}$ | AB | **0.004361206** | 0.051274735 | 0.020521847 | 0.006247895 | 0.008147895 |
|  | MB | 0.004419241 | 0.051059414 | 0.020803912 | 0.004361206 | 0.003454528 |
|  | SD | 0.004078875 | 0.016617355 | 0.021677285 | 8.7338E-15 | 6.37516E-15 |
| $f_{13}$ | AB | **-3.862782148** | -3.207627571 | -3.122812884 | -3.200286885 | -3.311538343 |
|  | MB | -3.862782148 | -3.222983851 | -3.198877457 | -3.200286885 | -3.615938695 |
|  | SD | 2.40793E-15 | 0.032397257 | 0.357126056 | 2.22045E-15 | 0.128463953 |
| $f_{14}$ | AB | **0** | 0.00060678 | 1.07786E-11 | 4.45378E-31 | 0.036347329 |
|  | MB | 3.82624E-12 | 0.000606077 | 0 | 4.93038E-32 | 0.002324632 |
|  | SD | 2.93547E-11 | 0.000179458 | 0 | 1.0696E-30 | 0.032374213 |

**Table 5.** Minimization results of benchmark functions in Table A3 with $n=30$. Maximum number of iterations=500.

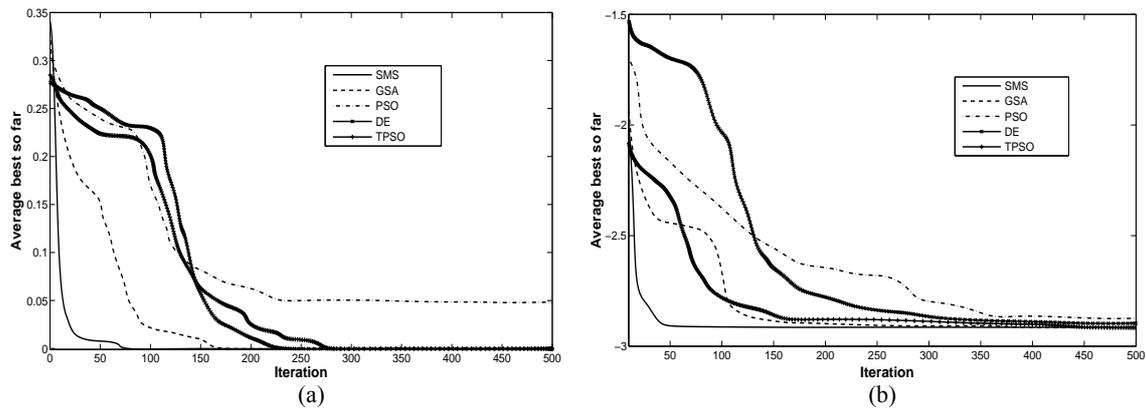

**Fig. 6.** Convergence rate comparison of PSO, GSA, DE, SMS and TPSO for minimization of (a) $f_{12}$ and (b) $f_{13}$.

Test functions from the GECCO contest

The experimental set in Table A4 includes several representative functions that are used in the GECCO contest. Using such functions, the SMS algorithm is compared to GSA, PSO, DE and TPSO. The results have been averaged over 30 runs, reporting the performance indexes for each algorithm in Table 6.

Likewise, *p*-values of the Wilcoxon signed-rank test of 30 independent executions are listed in Table 7. According to results of Table 6, it is evident that SMS yields much better solutions than other methods. The Wilcoxon test results in Table 7 provide information to statistically demonstrate that SMS has performed better than PSO, GSA, DE and TPSO. Figure 7 presents the convergence rate for the GSA, PSO, DE, SMS and TPSO algorithms, considering functions $f_{17}$ to $f_{24}$.

| | | SMS | GSA | PSO | DE | TPSO |
|---|---|---|---|---|---|---|
| $f_{15}$ | AB | **-25.91760733** | 57.15411412 | 134.3191481 | 183.6659439 | -18.63859195 |
| | MB | -29.92042882 | 57.38647154 | 133.1673936 | 186.723035 | -21.73646913 |
| | SD | 23.85960437 | 14.20175945 | 68.4414947 | 38.0678428 | 12.54569285 |
| $f_{16}$ | AB | **-57.89720018** | -57.89605386 | -40.5715691 | -52.92227417 | -50.437455071 |
| | MB | -57.89733814 | -57.89616319 | -40.00561762 | -53.25902658 | -52.564574931 |
| | SD | 0.00077726 | 0.000841082 | 4.812411459 | 1.769678878 | 1.3446395342 |
| $f_{17}$ | AB | **184.7448285** | 186.1082051 | 7540.2406 | 186.6192165 | 190.43463434 |
| | MB | 184.7424982 | 186.0937327 | 4831.581816 | 186.6285041 | 188.43649638 |
| | SD | 0.180957032 | 0.149285212 | 7101.466992 | 0.208918841 | 2.4340683134 |
| $f_{18}$ | AB | **-449.9936552** | 2015.050538 | 18201.78495 | -435.2972206 | -410.37493561 |
| | MB | -449.994798 | 1741.613119 | 18532.32174 | -436.0279997 | -429.46295713 |
| | SD | 0.005537064 | 1389.619208 | 6325.379751 | 2.880379023 | 1.4538493855 |
| $f_{19}$ | AB | **1213.421542** | 22038.7467 | 30055.82961 | 43551.34835 | 1452.4364384 |
| | MB | -181.0028277 | 21908.86945 | 26882.92621 | 42286.55626 | 1401.7493617 |
| | SD | 4050.267293 | 1770.050492 | 18048.55578 | 7505.414378 | 532.36343411 |
| $f_{20}$ | AB | **26975.80614** | 66771.65533 | 44221.12187 | 58821.82993 | 29453.323822 |
| | MB | 24061.19301 | 65172.39992 | 44733.97226 | 60484.33588 | 28635.439023 |
| | SD | 10128.06919 | 12351.81976 | 16401.44428 | 9191.787618 | 4653.1269549 |
| $f_{21}$ | AB | **6526.690523** | 23440.26883 | 23297.93668 | 26279.82607 | 7412.5361303 |
| | MB | 5716.886785 | 23427.99207 | 22854.63384 | 26645.28551 | 7012.4634613 |
| | SD | 2670.569217 | 2778.292017 | 5157.063617 | 2726.609286 | 745.37485621 |
| $f_{22}$ | AB | **965.8899213** | 181742714.4 | 7385919478 | 284396.8728 | 1051.4348595 |
| | MB | 653.8161313 | 196616193.9 | 5789573763 | 287049.5324 | 1003.3448944 |
| | SD | 751.3821374 | 79542617.71 | 5799950322 | 66484.87261 | 894.43484589 |
| $f_{23}$ | AB | **18617.61336** | 30808.74384 | 444370.5566 | 429178.9416 | 20654.323956 |
| | MB | 10932.4606 | 28009.57647 | 425696.8169 | 418480.2092 | 19434.343851 |
| | SD | 18224.4141 | 17834.72979 | 145508.9625 | 59342.54534 | 473.45938567 |
| $f_{24}$ | AB | **910.002925** | 997.4123375 | 1026.555016 | 917.4176502 | 1017.3484548 |
| | MB | 910.0020976 | 999.1456735 | 1025.559417 | 917.3421337 | 993.34434754 |
| | SD | 0.004747964 | 19.08754967 | 57.01221298 | 0.456440816 | 45.343496836 |

**Table 6.** Minimization results of benchmark functions in Table A4 with *n*=30. Maximum number of iterations=1000.

| SMS vs | GSA | PSO | DE | TPSO |
|---|---|---|---|---|
| $f_{15}$ | 1,7344E-06 | 1,7344E-06 | 1,7344E-06 | 5,2334E-05 |
| $f_{16}$ | 9,7110E-05 | 1,7344E-06 | 1,7344E-06 | 3,1181E-05 |
| $f_{17}$ | 1,12654E-05 | 1,7344E-06 | 1,7344E-06 | 6.2292E-05 |
| $f_{18}$ | 1,7344E-06 | 1,7344E-06 | 1,7344E-06 | 1.8938E-05 |
| $f_{19}$ | 1,92092E-06 | 1,7344E-06 | 1,7344E-06 | 9.2757E-05 |
| $f_{20}$ | 1,7344E-06 | 9,7110E-05 | 2,1264E-06 | 8.3559E-05 |
| $f_{21}$ | 1,7344E-06 | 1,7344E-06 | 1,7344E-06 | 7.6302E-05 |
| $f_{22}$ | 1,7344E-06 | 1,7344E-06 | 1,7344E-06 | 6.4821E-05 |
| $f_{23}$ | 0,014795424 | 1,7344E-06 | 1,7344E-06 | 8.8351E-05 |
| $f_{24}$ | 1,7344E-06 | 1,7344E-06 | 1,7344E-06 | 9.9453E-05 |

**Table 7.** *p*-values produced by Wilcoxon's test that compare SMS vs. GSA, SMS vs. PSO, SMS vs. DE and SMS vs. TPSO, for the "average best-so-far" (AB) values from Table 6.

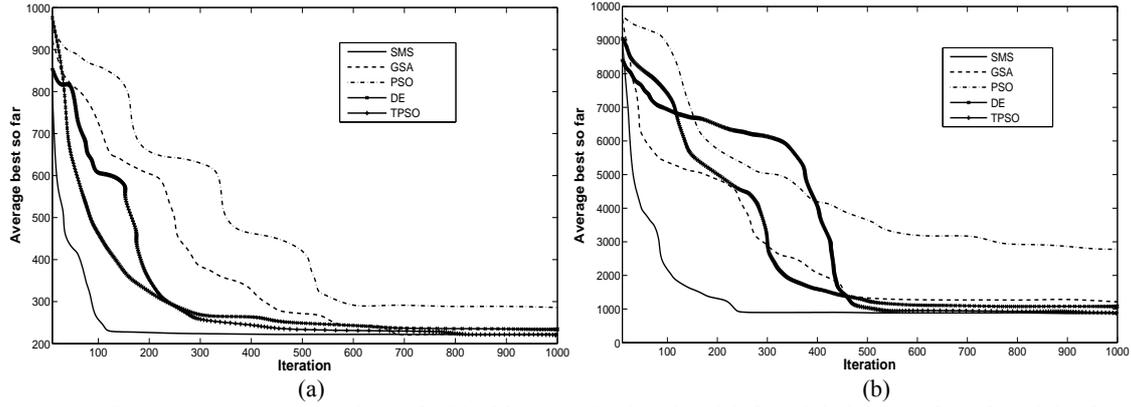

**Fig. 7.** Convergence rate comparison of PSO, GSA, DE, SMS and TPSO for minimization of (a) $f_{17}$ and (b) $f_{24}$.

## 5. Conclusions

In this paper, a novel nature-inspired algorithm called as the States of Matter Search (SMS) has been introduced. The SMS algorithm is based on the simulation of the State of Matter phenomenon. In SMS, individuals emulate molecules which interact to each other by using evolutionary operations that are based on physical principles of the thermal-energy motion mechanism. The algorithm is devised by considering each state of matter at one different exploration–exploitation ratio. The evolutionary process is divided into three phases which emulate the three states of matter: gas, liquid and solid. At each state, molecules (individuals) exhibit different movement capacities. Beginning from the gas state (pure exploration), the algorithm modifies the intensities of exploration and exploitation until the solid state (pure exploitation) is reached. As a result, the approach can substantially improve the balance between exploration–exploitation, yet preserving the good search capabilities of an EA approach.

SMS has been experimentally tested considering a suite of 24 benchmark functions. The performance of SMS has been also compared to the following evolutionary algorithms: the Particle Swarm Optimization method (PSO) [17], the Gravitational Search Algorithm (GSA) [16], the Differential Evolution (DE) algorithm [13] and the PSO algorithm with a territorial diversity-preserving scheme (TPSO) [39]. Results have confirmed a high performance of the proposed method in terms of the solution quality for solving most of benchmark functions.

The SMS's remarkable performance is associated with two different reasons: (i) the defined operators allow a better particle distribution in the search space, increasing the algorithm's ability to find the global optima; and (ii) the division of the evolution process at different stages, provides different rates between exploration and exploitation during the evolution process. At the beginning, pure exploration is favored at the gas state, then a mild transition between exploration and exploitation features during liquid state. Finally, pure exploitation is performed during the solid state.

**Appendix A. List of benchmark functions**

| Test function | $S$ | $f_{opt}$ | $n$ |
|---|---|---|---|
| $f_1(\mathbf{x}) = \sum_{i=1}^{n} x_i^2$ | $[-100,100]^n$ | 0 | 30 |
| $f_2(\mathbf{x}) = \max\{|x_i|, 1 \le i \le n\}$ | $[-100,100]^n$ | 0 | 30 |
| $f_3(\mathbf{x}) = \sum_{i=1}^{n-1} \left[ 100(x_{i+1} - x_i^2)^2 + (x_i - 1)^2 \right]$ | $[-30,30]^n$ | 0 | 30 |
| $f_4(\mathbf{x}) = \sum_{i=1}^{n} i x_i^4 + rand(0,1)$ | $[-1.28,1.28]^n$ | 0 | 30 |

**Table A1.** Unimodal test functions.

| Test function | S | $f_{opt}$ | n |
|---|---|---|---|
| $f_5(\mathbf{x}) = 418.9829n + \sum_{i=1}^{n}\left(-x_i \sin\left(\sqrt{|x_i|}\right)\right)$ | $[-500, 500]^n$ | 0 | 30 |
| $f_6(\mathbf{x}) = \sum_{i=1}^{50}\left(x_i^2 - 10\cos(2\pi x_i) + 10\right)$ | $[-5.12, 5.12]^n$ | 0 | 30 |
| $f_7(\mathbf{x}) = \frac{1}{4000}\sum_{i=1}^{n} x_i^2 - \prod_{i=1}^{n}\cos\left(\frac{x_i}{\sqrt{i}}\right) + 1$ | $[-600, 600]^n$ | 0 | 30 |
| $f_8(\mathbf{x}) = \frac{\pi}{n}\left\{10\sin(\pi y_1) + \sum_{i=1}^{n-1}(y_i - 1)^2\left[1 + 10\sin^2(\pi y_{i+1})\right] + (y_n - 1)^2\right\} + \sum_{i=1}^{n}u(x_i, 10, 100, 4)$ | $[-50, 50]^n$ | 0 | 30 |
| $y_i = 1 + \frac{(x_i + 1)}{4}$ $\quad u(x_i, a, k, m) = \begin{cases} k(x_i - a)^m & x_i > a \\ 0 & -a \leq x_i \leq a \\ k(-x_i - a)^m & x_i < a \end{cases}$ | | | |
| $f_9(\mathbf{x}) = 0.1\left\{\sin^2(3\pi x_1) + \sum_{i=1}^{n}(x_i - 1)^2\left[1 + \sin^2(3\pi x_i + 1)\right] + (x_n - 1)^2\left[1 + \sin^2(2\pi x_n)\right]\right\}$ $+ \sum_{i=1}^{n}u(x_i, 5, 100, 4)$ where $u(x_i, a, k, m)$ is the same as $f_8$ | $[-50, 50]^n$ | 0 | 30 |
| $f_{10}(\mathbf{x}) = \sum_{i=1}^{n}x_i^2 + \left(\sum_{i=1}^{n}0.5ix_i\right)^2 + \left(\sum_{i=1}^{n}0.5ix_i\right)^4$ | $[-10, 10]^n$ | 0 | 30 |
| $f_{11}(\mathbf{x}) = 1 - \cos(2\pi\|x\|) + 0.1\|x\|$ where $\|x\| = \sqrt{\sum_{j=1}^{n}x_j^2}$ | $[-100, 100]^n$ | 0 | 30 |

**Table A2.** Multimodal test functions.

| Test function | S | $f_{opt}$ | n |
|---|---|---|---|
| $f_{12}(\mathbf{x}) = \sum_{i=1}^{11}\left[a_i - \frac{x_i(b_i^2 + b_i x_2)}{b_i^2 + b_i x_3 + x_4}\right]^2$ $\mathbf{a} = [0.1957, 0.1947, 0.1735, 0.1600, 0.0844, 0.0627, 0.456, 0.0342, 0.0323, 0.0235, 0.0246]$ $\mathbf{b} = [0.25, 0.5, 1, 2, 4, 6, 8, 10, 12, 14, 16]$ | $[-5, 5]^n$ | 0.00030 | 4 |
| $f_{13}(\mathbf{x}) = \sum_{i=1}^{4}c_i \exp\left(-\sum_{j=1}^{3}A_{ij}(x_j - P_{ij})^2\right)$ $\mathbf{A} = \begin{bmatrix} 10 & 3 & 17 & 3.5 & 1.7 & 8 \\ 0.05 & 10 & 17 & 0.1 & 8 & 14 \\ 3 & 3.5 & 17 & 10 & 17 & 8 \\ 17 & 8 & 0.05 & 10 & 0.1 & 14 \end{bmatrix}$ $\mathbf{c} = [1, 1.2, 3, 3.2]$ $\mathbf{P} = \begin{bmatrix} 0.131 & 0.169 & 0.556 & 0.012 & 0.828 & 0.588 \\ 0.232 & 0.413 & 0.830 & 0.373 & 0.100 & 0.999 \\ 0.234 & 0.141 & 0.352 & 0.288 & 0.304 & 0.665 \\ 0.404 & 0.882 & 0.873 & 0.574 & 0.109 & 0.038 \end{bmatrix}$ | $[0, 1]^n$ | -3.32 | 6 |
| $f_{14}(\mathbf{x}) = (1.5 - x_1(1 - x_2))^2 + (2.25 - x_1(1 - x_2))^2 + (2.625 - x_1(1 - x_2))^2$ | $[-4.5, 4.5]^n$ | 0 | 2 |

**Table A3.** Multimodal test functions with fixed dimensions

 

| Test function | $S$ | $n$ | GECCO classification |
|---|---|---|---|
| $f_{15}(\mathbf{x}) = 10^6 \cdot z_1^2 + \sum_{i=2}^{n} z_i + f_{opt}$ <br> $\mathbf{z} = T_{osz}(\mathbf{x} - \mathbf{x}^{opt})$ <br> $T_{osz}: \mathbb{R}^n \to \mathbb{R}^n$, for any positive integer $n$, it maps element-wise. <br> $\mathbf{a} = T_{osz}(\mathbf{h})$, $\mathbf{a} = \{a_1, a_2, \ldots, a_n\}$, $\mathbf{h} = \{h_1, h_2, \ldots, h_n\}$ <br> $a_i = \text{sign}(h_i)\exp(K + 0.049(\sin(c_1 K) + \sin(c_2 K)))$, <br> where <br> $K = \begin{cases} \log(h_i) & \text{if } h_i \neq 0 \\ 0 & \text{otherwise} \end{cases}$, $\text{sign}(h_i) = \begin{cases} -1 & \text{if } h_i < 0 \\ 0 & \text{if } h_i = 0, \\ 1 & \text{if } h_i > 0 \end{cases}$ <br> $c_1 = \begin{cases} 10 & \text{if } h_i > 0 \\ 5.5 & \text{otherwise} \end{cases}$ and $c_2 = \begin{cases} 7.9 & \text{if } h_i > 0 \\ 3.1 & \text{otherwise} \end{cases}$ | $[-5,5]^n$ | 30 | GECCO2010 Discus function $f_{11}(\mathbf{x})$ |
| $f_{16}(\mathbf{x}) = \sqrt{\sum_{i=1}^{n} |z_i|^{2+4\frac{i-1}{n-1}}} + f_{opt}$ <br> $\mathbf{z} = \mathbf{x} - \mathbf{x}^{opt}$ | $[-5,5]^n$ | 30 | GECCO2010 Different Powers function $f_{14}(\mathbf{x})$ |
| $f_{17}(\mathbf{x}) = -\frac{1}{n}\sum_{i=1}^{n} z_i \sin(\sqrt{|z_i|}) + 4.189828872724339 + 100 f_{pen}\left(\frac{\mathbf{z}}{100}\right) + f_{opt}$ <br> $\hat{\mathbf{x}} = 2 \times \mathbf{1}_-^+ \otimes \mathbf{x}$ <br> $\hat{z}_1 = \hat{x}_1, \hat{z}_{i+1} = \hat{x}_{i+1} + 0.25(\hat{x}_i - x_i^{opt})$ for $i = 1, \ldots, n-1$ <br> $\mathbf{z} = 100\left(\Lambda^{10}(\hat{\mathbf{z}} - \mathbf{x}^{opt}) + \mathbf{x}^{opt}\right)$ <br> $f_{pen}: \mathbb{R}^n \to \mathbb{R}$, <br> $a = f_{pen}(\mathbf{h})$, $\mathbf{h} = \{h_1, h_2, \ldots, h_n\}$ <br> $a = 100\sum_{i=1}^{n} \max(0, |h_i| - 5)^2$ <br> $\mathbf{1}_-^+$ is a $n$-dimensional vector with elements of -1 or 1 computed considering equal probability. | $[-5,5]^n$ | 30 | GECCO2010 Schwefel function $f_{20}(\mathbf{x})$ |
| $f_{18}(\mathbf{x}) = \sum_{i=1}^{n} z_i^2 - 450$ <br> $\mathbf{z} = \mathbf{x} - \mathbf{x}^{opt}$ | $[-100,100]^n$ | 30 | GECCO2005 Shifted Sphere Function $f_1(\mathbf{x})$ |
| $f_{19}(\mathbf{x}) = \sum_{i=1}^{n}\left(\sum_{j=1}^{i} z_j\right)^2 - 450$ <br> $\mathbf{z} = \mathbf{x} - \mathbf{x}^{opt}$ | $[-100,100]^n$ | 30 | GECCO2005 Shifted Schwefel's Problem $f_2(\mathbf{x})$ |
| $f_{20}(\mathbf{x}) = \left(\sum_{i=1}^{n}\left(\sum_{j=1}^{i} z_j\right)^2\right) \cdot (1 + 0.4|N(0,1)|) - 450$ <br> $\mathbf{z} = \mathbf{x} - \mathbf{x}^{opt}$ | $[-100,100]^n$ | 30 | GECCO2005 Shifted Schwefel's Problem 1.2 with Noise in Fitness $f_4(\mathbf{x})$ |
| $f_{21}(\mathbf{x}) = \max\{|\mathbf{Ax} - \mathbf{b}_i|\} - 310$ <br> $\mathbf{A}$ is a $n \times n$ matrix, $a_{i,j}$ are integer random numbers in the | $[-100,100]^n$ | 30 | GECCO2005 Schwefel's Problem 2.6 with |

range $[-500,500]$, $\det(\mathbf{A}) \neq 0$.

$\mathbf{b}_i = \mathbf{A}_i \cdot \mathbf{o}$

$\mathbf{A}_i$ is the $i$-th row of $\mathbf{A}$ whereas $\mathbf{o}$ is a $n \times 1$ vector whose elements are random numbers in the range [-100 100].

Global Optimum on Bounds $f_5(\mathbf{x})$

$$f_{22}(\mathbf{x}) = \sum_{i=1}^{n}\left(100\left(z_i^2 - z_{i+1}\right)^2 + (z_i - 1)^2\right) + 390$$

$\mathbf{z} = \mathbf{x} - \mathbf{x}^{opt}$

$[-100,100]^n$ — 30 — GECCO2005 Shifted Rosenbrock's Function $f_6(\mathbf{x})$

$$f_{23}(\mathbf{x}) = \sum_{i=1}^{n}\left(A_i - B_i(\mathbf{x})\right)^2 - 460$$

$A_i = \sum_{j=1}^{n}\left(a_{i,j}\sin\alpha_j + b_{i,j}\cos\alpha_j\right)$  $B_i(\mathbf{x}) = \sum_{i=1}^{n}\left(a_{i,j}\sin x_j + b_{i,j}\cos x_j\right)$

For $i = 1,...,n$.

$a_{i,j}$ and $b_{i,j}$ are integer random numbers in the range [-100,100],

$\boldsymbol{\alpha} = [\alpha_1, \alpha_2, ..., \alpha_n]$, $\alpha_j$ are random numbers in the range $[-\pi, \pi]$.

$[-\pi, \pi]^n$ — 30 — GECCO2005 Schwefel's Problem 2.13 $f_{12}(\mathbf{x})$

$$f_{24}(\mathbf{x}) = \sum_{i=1}^{10}\hat{F}_i(\mathbf{x} - \mathbf{x}_i^{opt})/\lambda_i$$

$F_{1-2}(\mathbf{x}) =$ Ackley's function

$F_i(\mathbf{x}) = -20\exp\left(-0.2\sqrt{\frac{1}{D}\sum_{i=1}^{n}x_i^2}\right) - \exp\left(\frac{1}{D}\sum_{i=1}^{n}\cos(2\pi x_i)\right) + 20$

$F_{3-4}(\mathbf{x}) =$ Rastringin's function

$F_i(\mathbf{x}) = \sum_{i=1}^{n}\left(x_i^2 - 10\cos(2\pi x_i) + 10\right)$

$F_{5-6}(\mathbf{x}) =$ Sphere function

$F_i(\mathbf{x}) = \sum_{i=1}^{n}x_i^2$

$F_{7-8}(\mathbf{x}) =$ Weierstrass function

$F_i(x) = \sum_{i=1}^{n}\left(\sum_{k=0}^{k\max}\left[a^k\cos(2\pi b^k(x_i + 0.5))\right]\right) - n\sum_{k=0}^{k\max}\left[a^k\cos(2\pi b^k(x_i \cdot 0.5))\right]$

$F_{9-10}(\mathbf{x}) =$ Griewank's function

$F_i(\mathbf{x}) = \sum_{i=1}^{n}\frac{x_i^2}{4000} - \prod_{i=1}^{n}\cos\left(\frac{x_i}{\sqrt{i}}\right) + 1$

$\hat{F}_i(\mathbf{z}) = F_i(\mathbf{z})/F_i^{\max}$. $F_i^{\max}$ is the maximum value of the particular function $i$.

$\boldsymbol{\lambda} = \left[\frac{10}{32}, \frac{5}{32}, 2, 1, \frac{10}{100}, \frac{5}{100}, 20, 10, \frac{10}{60}, \frac{5}{60}\right]$

$[-5,5]^n$ — 30 — GECCO2005 Rotated Version of Hybrid Composition Function $f_{16}(\mathbf{x})$

The $\mathbf{x}^{opt}$ and $f_{opt}$ values have been set to default values which have been obtained from the Matlab© implementation for GECCO competitions, as it is provided in [51].

**Table A4.** Set of representative GECCO functions.